# A Review on Drivers' Red Light Running Behaviour Predictions and Technology Based Countermeasures

Md Mostafizur Rahman Komol[1,2], Jack Pinnow[3], Mohammed Elhenawy[1,2], Shamsunnahar Yasmin[1,2], Mahmoud Masoud[1,2], Sebastien Glaser[1,2], Andry Rakotonirainy[1,2]
[1] Queensland University of Technology, Centre for Accident Research and Road Safety-Queensland, Australia
[2] Queensland University of Technology (QUT), Institute of Health and Biomedical Innovation (IHBI), Australia
[3] Department of Transport and Main Road (Queensland), Australia

Corresponding author: Md Mostafizur Rahman Komol (e-mail: mdmostafizurrahman.komol@hdr.qut.edu.au).

**ABSTRACT** Red light running at signalised intersections is a growing road safety issue worldwide, leading to the rapid development of advanced intelligent transportation technologies and countermeasures. However, existing studies have yet to summarise and present the effect of these technology-based innovations in improving safety. This paper represents a comprehensive review of red-light running behaviour prediction methodologies and technology-based countermeasures. Specifically, the major focus of this study is to provide a comprehensive review on two streams of literature targeting red-light running and stop-and-go behaviour at signalised intersection – (1) studies focusing on modelling and predicting the red-light running and stop-and-go related driver behaviour and (2) studies focusing on the effectiveness of different technology-based countermeasures which combat such unsafe behaviour. The study provides a systematic guide to assist researchers and stakeholders in understanding how to best identify red-light running and stop-and-go associated driving behaviour and subsequently implement countermeasures to combat such risky behaviour and improve the associated safety.

## I. INTRODUCTION

### A. BACKGROUND

Road crashes and the resulting fatalities are one of the major factors contributing to approximately 1.3 million people dying annually from road traffic crashes globally. A further 20 to 50 million suffer from life-threatening injuries as a result of motor-vehicle related crashes [1, 2]. Although some countries with a greater income level per capita have been able to achieve a significant reduction in road crash-related fatalities, the global rate is still alarmingly high, with 18.2 deaths per 100,000 population [2]. Intersection crashes represent a significant portion of these fatalities and life-threatening injuries, with the USA, for instance, reporting 45% of severe injuries and 22% of fatal crashes occurring at intersections [3]. The major factors of signalised intersection crashes are identified to be a failure to break immediately at the red light onset, miscalculating turning angles and distances of other vehicles, and deliberately running the red light [4]. At signalised intersections, drivers are often faced with such judgement errors during the onset of yellow or amber signals and may become confused and accelerate through the amber light or brake hard, potentially resulting in a rear end or right angle collision and or serious injury outcomes [5, 6]. Drivers may also commit a red light running (RLR) violation due to misjudgment at the onset of yellow at a signalised intersection.

Qian and Dong [7] defined the RLR behaviour as the act of a vehicle encroaching the intersection once a red signal has appeared. Combating the RLR violation is a major safety concern due to the associated high crash risk and the crash severity outcomes resulting from such unsafe driving behaviour [8, 9]. Transport authorities around the world invest significant efforts to devise measures to reduce this unsafe behaviour and the associated safety risks. For instance, in the USA during 2005, a $14 billion investment served to identify different intervention strategies [10].

The RLR behaviour of drivers can be categorised as either passive or deliberate factors. Passive factors can be defined as drivers' inadvertent decisions caused by insensitivity and surrounding events, including driving inexperience, age, weather, road geometry and obstructed views. Deliberate factors are the deliberate decisions of a driver to be engaged in dangerous driving behaviour, which include an aggressive driving nature, peer pressure, and speeding to save time [11-13]. RLR associated with such deliberate factors is likely to result in frequent crashes with a higher level of injury severity outcomes at signalised intersections [14].

As such, driver behaviour is identified as one of the most critical factors in devising countermeasures to combat RLR behaviour [15], receiving significant attention from road safety researchers. Researchers have developed machine learning models such as random forest, support vector machine, artificial and deep neural networks, and econometric models. The use of advanced artificial intelligence (A.I.) techniques, such as machine learning (ML) and deep learning (DL) algorithms, has added significant value to the modelling and prediction of driver behaviour [16] since they are able to incorporate driver intention measures, driving manoeuvres, driver characteristics, vehicle and driver state, weather conditions, environment and light conditions as exogenous features. More recently, understanding the effect of advanced driving assistant systems (ADAS) on driver behaviour has received significant attention [17].

*B. CONTRIBUTIONS*

The major focus of this study is to provide a comprehensive literature review focusing on driver behaviour prediction models specific to RLR and stop-go decisions in the dilemma zone. While previous literature analysed similar topics of RLR and stop-go decisions at yellow onset [17, 18], the studies that review the effect of advancement in vehicle safety features and their impact on RLR behaviour are few. At the same time, the significant advancements in modelling and data collection techniques have enabled better and more accurate prediction of RLR behaviour, with technology-based countermeasures targeting RLR behaviour also improving considerably over the last five years. With the emergence of new technology such as ADAS and connected and automated vehicles (CAV), it is of utmost importance to revisit the effect of these countermeasures and understand how these may impact RLR behaviour. Therefore, this comprehensive review is of the two streams of literature targeting RLR and stop-and-go (SAG) behaviour – (1) studies focusing on modelling the RLR and SAG related driver behaviour, and (2) studies focusing on the effectiveness of different technology-based countermeasures in combating unsafe behaviour. The first stream of studies evaluates the modelling techniques used to predict RLR and SAG related driving behaviour, the associated model accuracy measures, the parameters considered in these models, along the methods adopted for collecting data for such models. The second stream of studies served to analyse the progress in devising the countermeasure technologies and innovations targeted toward combating unsafe driving behaviours. Together, these analyses provide a systematic guide to enable researchers to best identify RLR and SAG associated driving behaviour and subsequently implement countermeasures to combat such risky behaviour and improve safety.

The rest of the paper is organised as follows. In the first section, the methodology and research protocol to identify the most relevant studies focusing on the SAG and RLR prediction are presented. In the second section, the causes and consequences of SAG and RLR behaviour are discussed, along with a succinct discussion of typical modelling techniques and parameters. In the third section, the models selected are then presented and discussed in detail. In the fourth section, current countermeasure technologies are identified and reviewed. Lastly, future recommendations are put forward and conclusions are drawn.

## II. METHODOLOGY AND RESEARCH PROTOCOL

Titles, abstracts, and keywords of existing relevant studies were extracted from electronic databases into EndNote X9. Screening of studies was conducted in three stages. First, duplicates were removed using an in-built EndNote function, followed by manual inspection and removal of remaining duplicates. Following this, titles and abstracts were screened for papers highly unrelated to the topic of the review and removed, followed by a second, more focused screening and removal process.

A search strategy was employed to identify all relevant articles for the systematic review until May 2021 (shown in Fig 1). The identification stage occurred using a keyword search strategy in Scopus, Web of Science and IEEE electronic databases. This search utilised the in-built boolean functions in the electronic databases. The keyword string consisted of three primary segments, which were all joined with the 'AND' boolean. The first segment targeted all papers relevant to red light running and dilemma zones by joining the following keywords with the 'OR' boolean; "Red light running", "Red light turning", "Dilemma zone", "Dilemma behaviour", and "RLR". All models were identified in the second-string segment by searching for "Model", "Prediction", "Countermeasure", and "Behaviour". Lastly, to remove any papers unrelated to the field of car safety, "Vehicle" was also added at the end of the search. After importing all relevant papers into the bibliographic software, EndNote, papers were manually excluded based on their relevance. Irrelevant articles included those which focused on predicting, analysing or offering countermeasures for motorcycles and pedestrians only, rather than conventional vehicles. Articles that predicted pedestrian behaviour in vehicle-pedestrian interactions were also excluded, as well as any papers related to policy measures. The systematic review scheme is shown in following Figure 1.

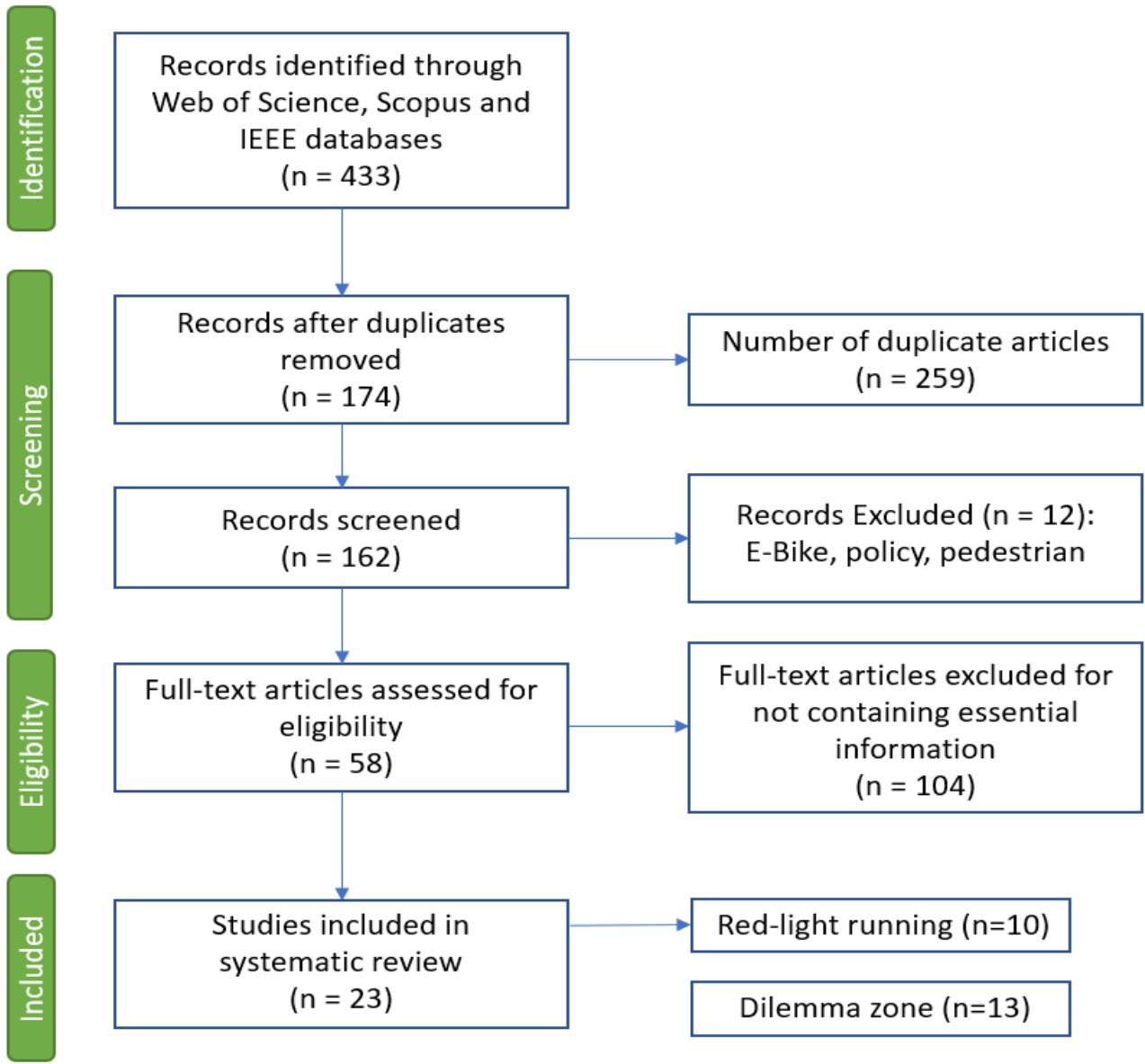

FIGURE 1. Systematic review scheme

The eligibility criteria for choosing RLR and SAG papers in our review were dependent on the article being assessed. RLR-based articles required the inclusion of prediction methods. Articles that focused on predicting SAG decisions at the yellow onset were also reviewed in the chapter.

A total of 58 papers were identified using the above-mentioned eligibility criterion. Each article was read fully, and the data extracted included the data source, predictor variables, prediction model (with any parameters) and model accuracy. During the manual data extraction, 3 papers were removed because the full-text version could not be accessed by the authors, and 2 papers were removed because they were conference proceedings that were expanded and improved further in published peer-reviewed articles by the same authors later. Many papers which focused on RLR or SAG decisions did not predict the act of RLR or stop-go, but rather a feature that was implicit in the act. Papers that focused on using RLR or stop-go to predict distraction, aggression or other behaviours were also excluded. These ineligibility criteria removed 35 papers, leaving 23 relevant papers (10 RLR, 13 SAG) for further analysis. The results of the review on RLR and SAG behaviour prediction are demonstrated in Table 1 and 2. Unless otherwise specified, the model accuracy extracted was the accuracy metric for the best performing prediction model presented in the associated paper. An exception to this is the work of Zhou, et al. [8], where the accuracy reported is the best mean accuracy of 6 models.

**Table 1: RLR Prediction Models**

| Author | Variables | Data source | Model | Accuracy |
|---|---|---|---|---|
| Chen, et al. [18] | Vehicle trajectory, position in the platoon | Radar sensor | Bayesian Network | 96.3% |
| | | | SVM (linear) | - |
| | | | SVM (non-linear) | |
| | | | Random Forest | |
| | | | Logistic Regression | |
| Gazis, et al. [5] | Vehicle trajectory, signal | Video footage | Nested logit model | - |
| Jahangiri, et al. [19] | Vehicle trajectory signals, epidemiological, distraction | Simulator, radar, video footage, signal | Random Forest | - |
| Jahangiri, et al. [20] | Vehicle trajectory | Radar, video, signal | SVM (Gaussian kernel) | 97.9% |
| | | | R.F. (500 trees, Gini criteria) | 93.6% |
| Li, et al. [21] | Vehicle trajectory vehicles queued, distance to head vehicle | Simulation | ANN | 83% |
| Pugh and Park [22] | Vehicle trajectory Epidemiological, distraction, dilemma zone type, pedal direction change | Experiment, Video footage | ANN | 82.04% |
| Ren, et al. [23] | Vehicle occupancy, the time gap between vehicles, signal | Loop detectors | Modified rare events logistic regression model | 77.3% |
| | | | Binary logistic model | 49.2% |
| Wang, et al. [24] | Vehicle trajectory, distance to stop line, headway between vehicles | field experiment, video sensors | Probabilistic model | 89.8% |
| Zaheri and Abbas [25] | Vehicle trajectory, signal | Radar data with signal phase, detector, and video data | Discriminant analysis algorithm | 96% |
| Zhang, et al. [26] | Vehicle trajectory, signal | Video footage | DARE-based Probabilistic model | >70% |

**Table 2: Stop-Go prediction models at yellow onset**

| Author | Variables | Data source | Model | Accuracy |
|---|---|---|---|---|
| Bar-Gera, et al. [27] | Vehicle trajectory and type, Distance to stop line | Digital cameras (naturalistic) | Four parameters logistic regression | - |
| | | | Exponential logistic regression | |
| Biswas and Ghosh [28] | Vehicle trajectory, signals, distance to stop line | Video footage, radar | Logistic Regression | 81.5% |
| | | | Fuzzy Logic | 87% |
| | | | ANN | 88.7% |
| | | | Weighted Average Hybrid Model | 96.2% |
| Ding, et al. [29] | Signals, occupancy, the distance between vehicles, vehicle decisions in the platoon and adjacent lane | Event-based vehicle-detector actuation and signal phase change (SMART-SIGNAL), stop bar detectors | Gradient boosting logic model | 95.2% |
| | | | Binary logit regression model | 75% |

| Author | Variables | Data source | Model | Accuracy |
|---|---|---|---|---|
| Pawar, et al. [30] | Road geometry, lane number, speed, distance, time, vehicle type : (car, truck, motorised two wheeler) | Video footage | Probabilistic model | 90.8% |
| Elhenawy, et al. [31] | Vehicle trajectory, epidemiological, pavement condition, driver behaviour | Onboard data acquisition system | Logistic Regression AdaBoost Random Forest SVM | 84.2% 83% 82.8% 90% |
| Li, et al. [3] | Vehicle trajectory, signal, position in the platoon, occupancy/queue length | Video footage | Hidden Markov Model | - |
| Ghanipoor Machiani and Abbas [32] | Vehicle trajectory, distance to stop line, pavement condition, position in the platoon, distraction | Driving simulator | Discriminant analysis | 99.4% |
| Pathivada and Perumal [33] | Vehicle trajectory and vehicle type: (motorised two wheeler, car, truck, motorised three wheeler), signals, distance to stop line, intersection type | Video footage | Binary logistic regression | 83.3% |
| Che Puan, et al. [34] | Vehicle trajectory and vehicle type: (passenger car, truck, bus, heavy truck, motorcycle), distance to stop line, position in the platoon | Video footage | Binary logistic regression | - |
| Tan, et al. [35] | Vehicle trajectory, distance to stop line | Video footage | Discrete choice model | - |
| Savolainen [36] | Distraction, epidemiological | Driving simulator | Random parameter logit Latent class logit | - |
| Wang, et al. [37] | Vehicle trajectory, signals | Video footage | Gazis Herman Maradudin Model | - |
| Zhou, et al. [8] | Physiological | EEG, simulator | BPNN | 91.4% |

## III. CAUSES AND CONSEQUENCES OF RLR AND SAG BEHAVIOUR

Understanding the psychology behind RLR and SAG behaviour is a necessary step to identify the critical parameters for use in predictive models. RLR and SAG are often associated with misjudgment at the onset of a yellow light, where the driver requires an immediate and decisive action during the approach to the intersection, known as the dilemma zone [38, 39]. Indecision in this zone can lead to the driver stopping or accelerating at an inappropriate time, resulting in a crash. Decisive actions at the dilemma zone during the amber signal may also result in RLR. Transportation researchers defined the dilemma zone as either type I or type II [39-42]. Dilemma zone type I is likely to occur due to inadequate intersection planning with an inaccurate sensor setup and signal timing [43, 44]. The Dilemma Zone at a four-armed signalised intersection is presented in Figure 2.

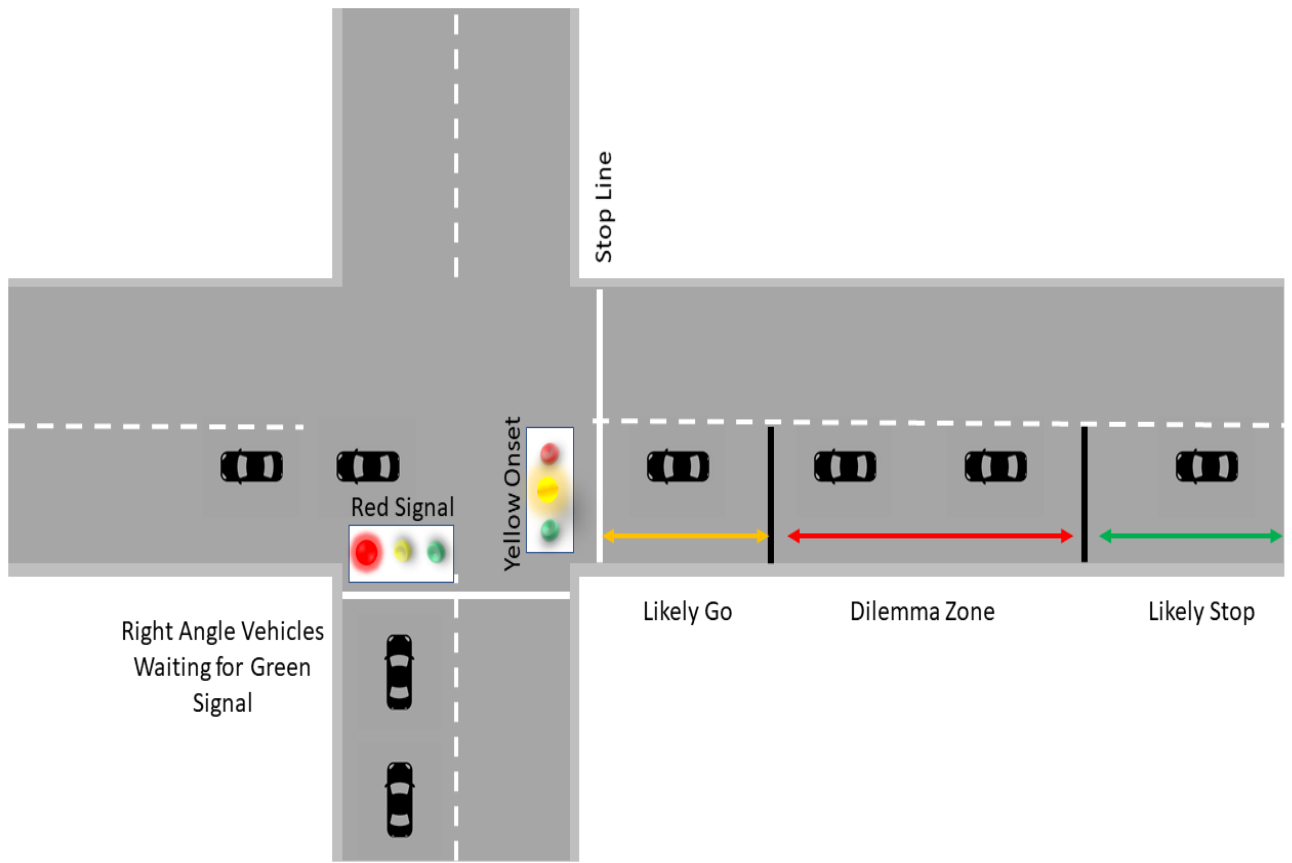

FIGURE 2. Dilemma zone detection in four-armed signalised intersection [Example is for the right-hand driving condition] [14, 38, 39] (https://icograms.com/icograms-designer.php).

In the dilemma zone, drivers are less likely to maintain the necessary distance to avoid collisions when instant breaking is required. This may be the result of the yellow signal turning red while the driver is within the intersection, potentially causing a right angle collision [5]. In dilemma zone type II scenarios, drivers are indecisive when actions are required [41], for example, when drivers stop at the onset of yellow but intend to run the intersection if they are near the stop line [45]. This indecision has been termed as the option zone, and it is unwise to underestimate this problem in road safety considerations [32]. The stochastic behaviour of drivers near the intersection, such as the type II dilemma zone, is seldom predictable, and such behaviours are likely to be associated with fatal crashes and serious injuries [38]. RLR at signalised intersections can lead to different types of crashes, but most commonly are rear end [46] and right angle collisions. A rear end crash is likely to occur from the sudden stopping decision by the driver of a lead vehicle at the yellow phase [47, 48], while a right angle crash can be the result of RLR at the yellow onset [5, 6, 49] Rear end collisions often result in minor injuries, such as whiplash [46], while right angle crashes are likely to result in higher injury severity outcomes due to the dissipation of higher kinetic energy of such crashes. Based on a Florida study in 2006, approximately 45% of right angle crashes resulted in incapacitating, non-incapacitating and fatal injuries [50]. Speeding and acceleration were found to be important factors in right angle collisions by Ogden and Newstead [51], noting that misunderstanding, driving inexperience, vehicle type and the number of lanes also contributed to right angle collision risk. Rear end (a) and right angle (b) collisions are shown in Figure 3.

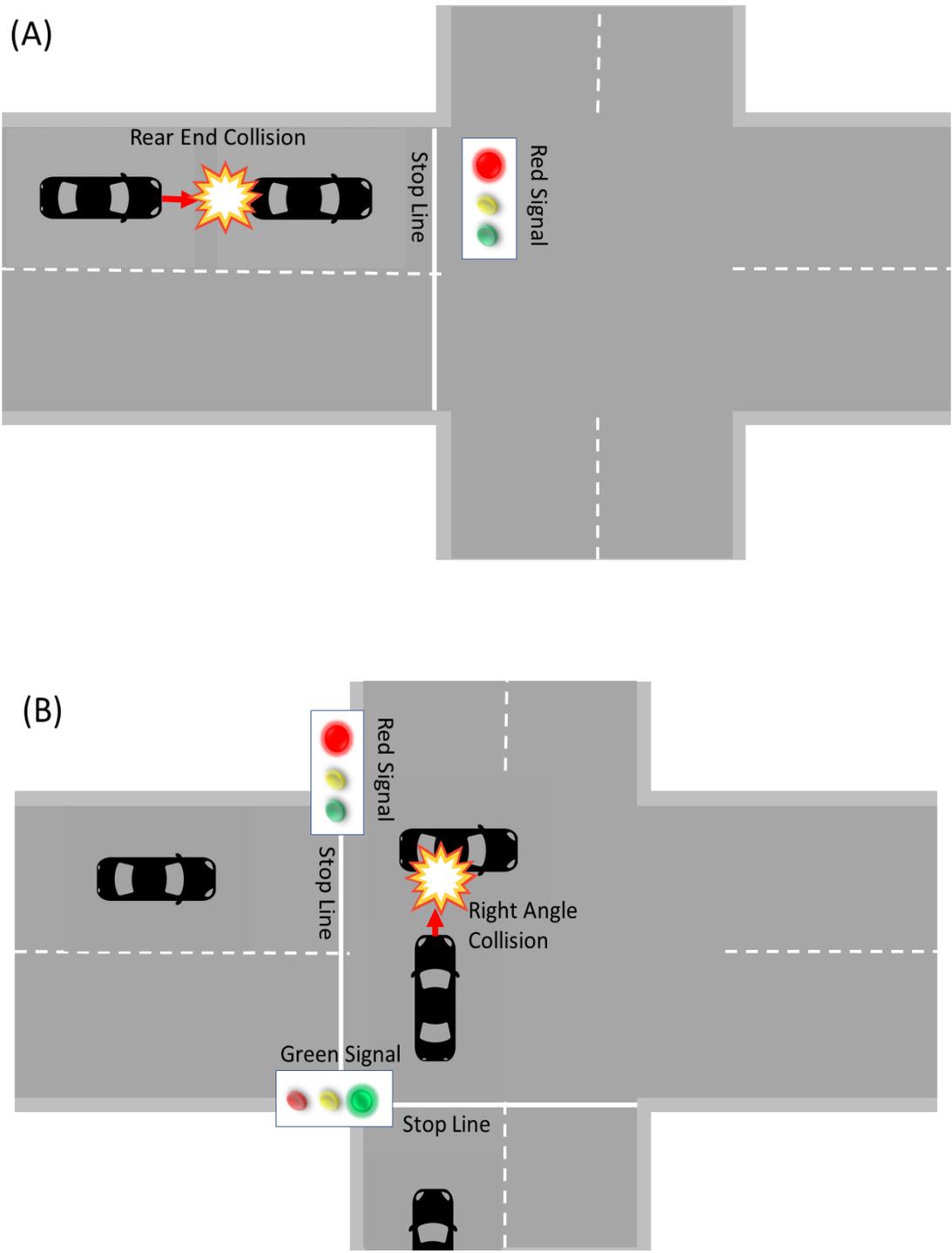

FIGURE 3. Traffic collisions in the dilemma zone due to red light running: (A) the rear end collision, (B) the right angle collision. (https://icograms.com/icograms-designer.php).

## IV. RESULT AND DISCUSSION ON PREDICTION OF RLR AND SAG BEHAVIOUR

Different driving features and human characteristics like reaction time, lane positioning, maneuvering, frequency of using the car horn, heart rate and eye movement have been noted to influence driver response at the intersection [52-54]. Differences in the driving experience, such as the difference between professional and non-professional drivers, also impact the likelihood of a driver engaging in RLR. For instance, taxi drivers have been shown to have a higher probability of engaging in RLR but have lower crash rates [55]. Some external distractions such as cell phone interfaces and phone calls have been identified as having a significant impact on driver behaviour at the dilemma zone [44]. The influence of weather on driver behaviour has also been shown to be significant [56]. Chen, et al. [57] mentioned that snow weather might increase the likelihood of engaging in RLR, but more data was necessary to confirm this claim. Some vehicle-related characteristics such as speed and pedal activity have been used to predict driver behaviour [58]. External prediction methods, such as video surveillance, have proven useful as a less intrusive method of measuring predictor variables [3]. More intrusive methods, such as electroencephalogram (EEG), can be used to give incredibly accurate information about drivers speeding behaviour which will help in SAG behaviour prediction [8] though this method is still considered too intrusive for commercial implementation.

Several different models of RLR and SAG behaviour prediction have been adopted which utilise the aforementioned variables. These models include the logistic regression model, Bayesian model, Probit model, decision tree model, agent-based model, and the fuzzy logic model incorporated with a classical logic model [55, 59-73]. Pathivada and Perumal [74] performed some research utilising numerous controlled experiments in India, including both two- and three-wheeler vehicles with cars and trucks to analyse mixed traffic conditions and employed binary logistic regression models from different experimental data to demonstrate driver behaviours at the dilemma zone. However, their experimental data did not include intersection geometry, surrounding land-use, signal visibility, driver knowledge on signal phases, type of vehicle, emotional state of the driver or distraction due to in-vehicle technology attributes [33, 38, 74]. The effect of surrounding vehicles on RLR at the dilemma zone has also incorporated the level of aggressiveness exhibited by drivers at the onset of the yellow signal. Pathivada and Perumal [38] used a statistical approach to SAG behaviour prediction, which analysed driver behaviour on homogeneous and mixed traffic conditions. Predictions of RLR have also been extended to connected vehicles in mixed traffic conditions by integrating connected vehicle data and traffic sensor information. From analysing Table 1 and Table 2, the number of RLR and SAG studies based on modelling technique and predictor variables are shown in Table 3 and the significance of these predictor models based on their use in number of models in different RLR and SAG studies is shown in Table 4.

Table 3: RLR and SAG studies for different modelling techniques and predictor variables

| No of Studies | | RLR | SAG |
|---|---|---|---|
| Modelling Techniques | Machine Learning | 2 | 2 |
| | Neural Network | 2 | 3 |
| | Regression | 2 | 6 |
| | Discriminant Analysis | 1 | 1 |
| | Probabilistic | 2 | 1 |
| | Other | 1 | 5 |
| Predictor Variables | Trajectory | 9 | 9 |
| | TTI | 4 | 8 |
| | DTI | 8 | 7 |
| | Other Vehicles | 5 | 5 |
| | Yellow Time | 4 | 6 |
| | Epidemiological | 1 | 3 |
| | Vehicle Type | 0 | 3 |
| | Other | 3 | 7 |
| Total Article Number Considered in the Study | | 10 | 13 |

Table 4: Significance of predictor variables

| Predictor Variables | Significance in Number of Models used in Selected SAG and RLR Articles | |
|---|---|---|
| | Significant | Considered But Not Significant |
| Trajectory | 8 | 4 |
| TTI | 6 | 5 |
| DTI | 6 | 3 |
| Other Vehicles | 5 | 4 |
| Yellow Time | 5 | 4 |
| Epidemiological | 3 | 2 |
| Vehicle Type | 3 | 0 |
| Other | 8 | 2 |

### A. DISCUSSION ON MODELS SELECTION AND PREDICTION ACCURACY

In predicting RLR and SAG behaviour, logit-based models have been the most common, followed by neural networks and random forest. The binary logit model is considered standard due to the binary decision of SAG or run versus do not run in the case of the RLR decision. When considering RLR prediction, regression has been adopted in fewer studies compared to stop-go decision studies. Several variations of this model have been proposed and are demonstrably better.

The accuracy of binary logistic regression models is typically found to be lower than other logit-based models, with the exception of work performed by Pathivada and Perumal [38]. Model accuracy is heavily influenced by the data collection technique. For instance, accuracy produced by loop detectors for binary logit models was inferior to the accuracy of SMART-SIGNAL detectors [29] and video recording and analysis [38]. Multinomial logistic regression models have proven to be accurate, with all accuracies reported at over 80%. The model types employed for examining RLR and SAG decisions are illustrated in Figure 4.

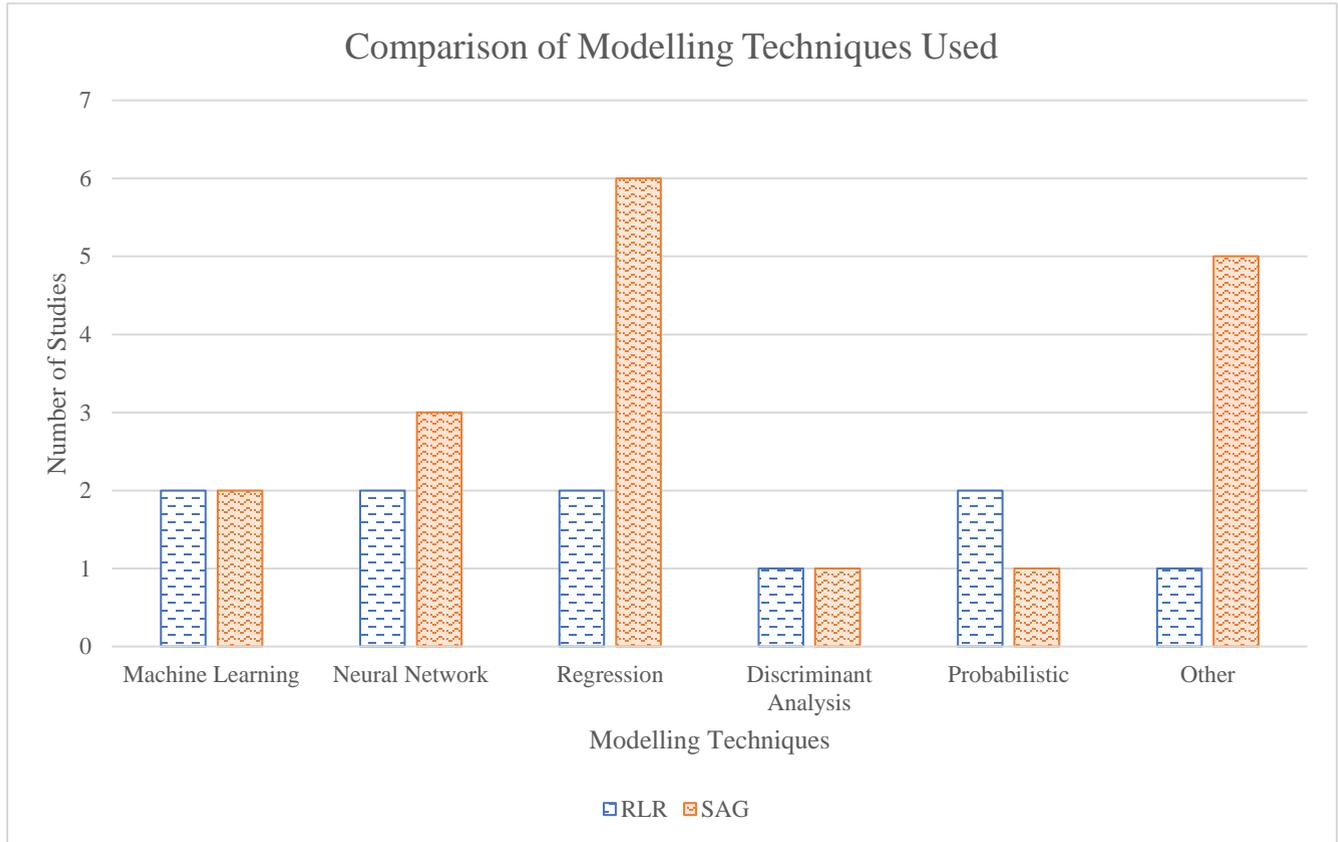

FIGURE 4. Comparison of model types found in the study for RLR and SAG behaviour

Extensions of these basic logistic regression models have been adopted by several authors [23]. The use of modified rare events logistic regression by Ren, et al. [23] was identified to provide a better fit in recognising RLR events. The model accuracy was considerably higher than the binary logistic model the authors tested. However, the modified rare event logistic regression model by Ren, et al. [23] showed reduced performance when compared with other model performance. The performance reduction of the rare event logistic regression model is likely due to the data source solely being based on data from a single loop detector 122m from the stop line. Random parameter and latent class logit models have been used to analyse signal timing strategies [36]. Based on the research findings by Savolainen [36], drivers were more partial to stop when warning flashes and enforcement cameras were present. Four parameter and exponential logistic regression models were used to estimate the actual entrance time of vehicles at an intersection [27]. Lower residuals were achieved for the exponential logistic regression model for most of the intersections considered in the analysis.

Machine learning models have also proven to be quite accurate in predicting RLR and SAG behaviour, with neural network-based models the most prominent, followed by random forest. The incorporation of simulation [21], video footage [22], radar [28] and EEG signals [8] as data sources to produce high accuracy neural network models demonstrates the robustness of this technique. However, the lack of models which incorporate lower resolution data, such as loop detectors, means they may be less appropriate than more rudimentary models in some circumstances. A weighted average hybrid model (WAHM) proposed by Biswas and Ghosh [28], which is a coupling of ANN and fuzzy logic models, was shown to have an incredibly high level of accuracy (96.15%) when detecting both crossing and stopping vehicles.

Random forest and AdaBoost models have been used with several different optimal hyperparameters for RLR behaviour prediction. Jahangiri, et al. [19] used 800 trees, with 6 factors considered per tree. Jahangiri, et al. [20] used 500 trees and 2 factors per tree. Conversely, Elhenawy, et al. [31] only used 50 trees in their study regarding the impact of driver aggression on RLR events. Findings from Elhenawy, et al. [31] and Jahangiri, et al. [20] showed that, in comparison to other models tested, the random forest was a poor performer in terms of SAG behaviour prediction that has reasonable, due likely to the tree-type structure of the model not being able to capture the nuances of a driver's decision to run a red light. Neither of these models considered leading vehicle behaviour, which is often an influential factor in drivers' SAG and RLR behaviour prediction.

Other machine learning models, such as SVM and discriminant analysis, have demonstrated very high levels of accuracy. SVM with a Gaussian kernel scored 90+% in studies where both were used. Similarly, discriminant analysis (DA) has shown to be the best performing model in terms of accuracy, scoring over 95% accuracy. However, the ability to adapt these DA models for realistic use is ambiguous, considering simulator studies are vastly different to observational studies, and the data sourcing and post-processing required to produce the inputs into these models is time-consuming. By contrast, SVM models have used lower resolution observational data and still maintained good levels of accuracy.

Probabilistic models have also been shown to have inherent advantages over their deterministic counterparts and have incorporated DARE into their formulations. The use of point detectors [26] and inductive loop detectors [24] in these models demonstrate their usefulness for SAG behaviour prediction, as these only require rudimentary technology, which has been more commonly implemented than other data capture devices (i.e. video footage followed by post-processing). Since there are only two probabilistic models with largely varying degrees of accuracy, it is difficult to know whether this method is accurate. The Bayesian network (B.N.) model used 17-day historical radar data to build the prediction model and achieve a high level of accuracy [18]. Notably, the B.N. model, which considered all features performed better than R.F., non-linear SVM and logistic regression. Importantly, the findings of the research by Chen, et al. [18] suggested that continuously monitored trajectory, used as a data source, produced a significantly better model than inductive loop detector data. A comparison of model accuracy and the data types used are illustrated in Fig 5.

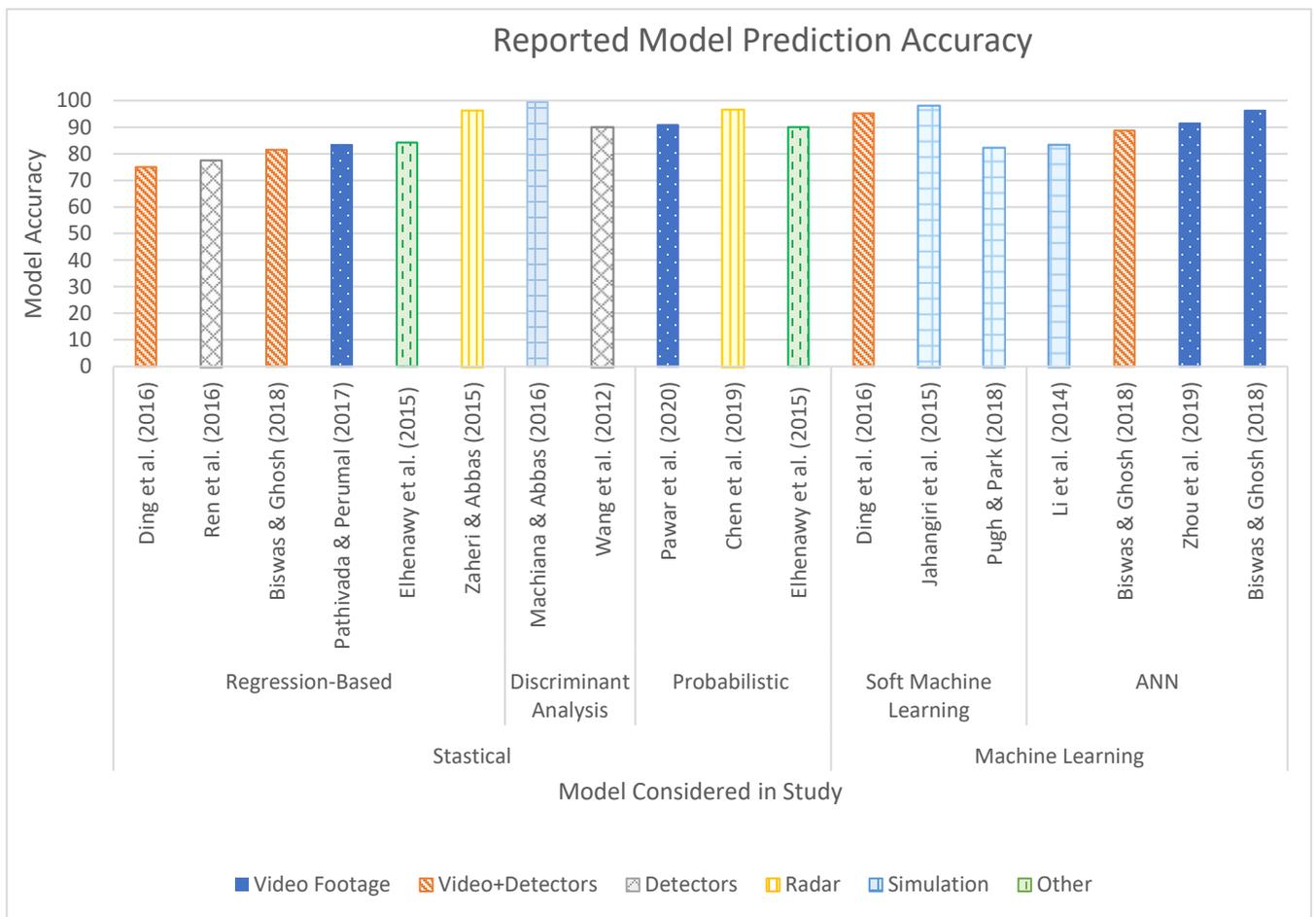

FIGURE 5. Comparison of reported model accuracy for different data collection methods

## B. PREDICTOR VARIABLES CONSIDERED IN STUDIES

Several predictor variables are consistently highlighted within the literature and used to identify RLR and SAG decisions made by drivers. Of these variables, time-to-intersection (TTI), distance-to-intersection (DTI), vehicle trajectory related variables, consideration of other vehicles, yellow time remaining, epidemiological and vehicle type are the most commonly found. Other variables identified within the literature (denoted 'Other') included driver aggression [31], pavement condition [31, 32], distraction [36], physiological signals [8] and police presence [32]. The frequency in which these variables appear within the literature reviewed is illustrated in Figure 6.

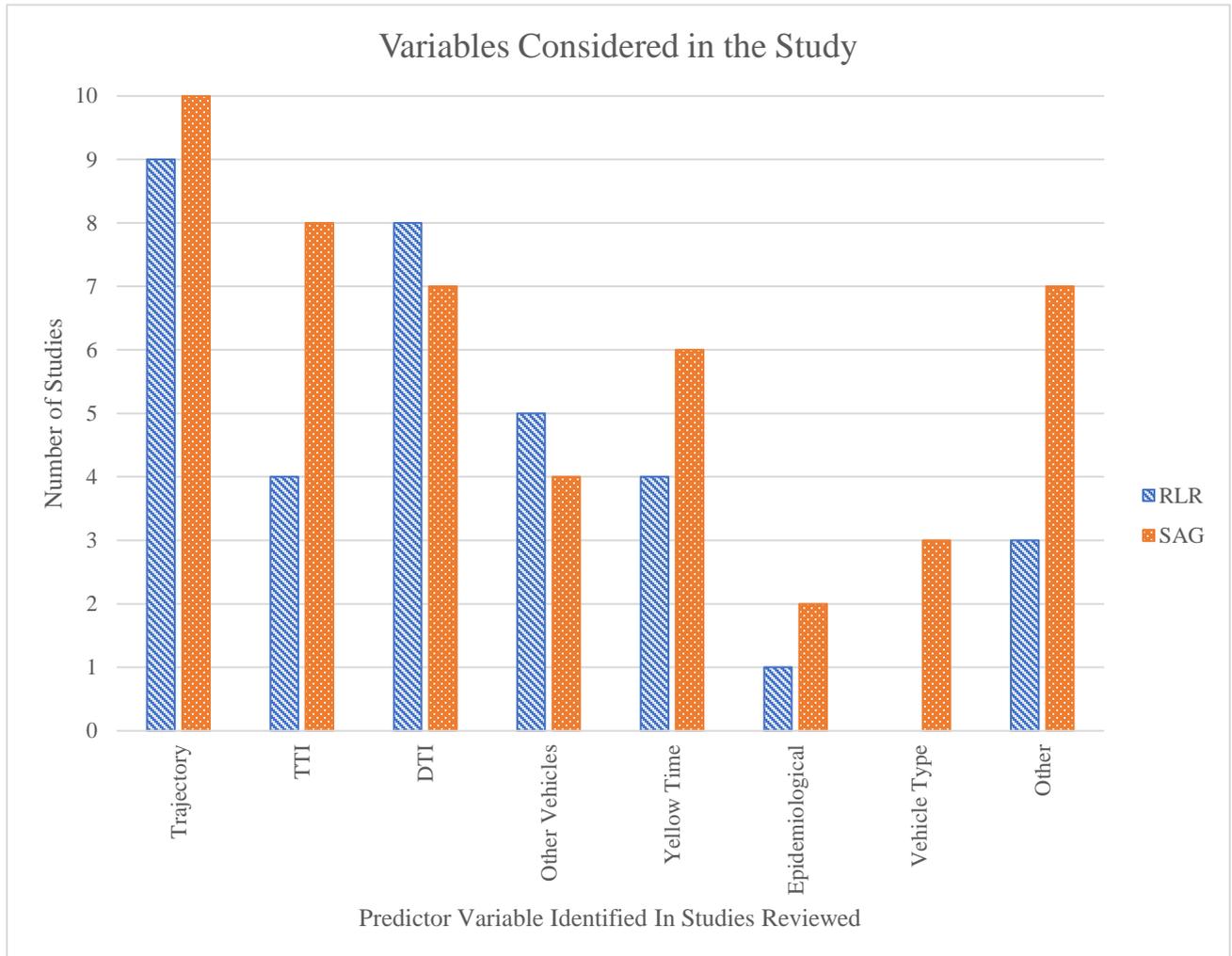

FIGURE 6. Common predictor variables identified in RLR and SAG studies.

Of these papers, few have indicated the most significant of all predictor variables considered. Variables were considered significant if the p-value in the regression was less than 0.05 in some cases [31, 34]. Other authors preferred a p-value of 0.01 to demonstrate significance [23, 29, 32, 33]. Gates, et al. [75] used model elasticity to identify which features were most indicative of RLR behaviour [75]. Moreover, feature importance from random forest models produced by Jahangiri, et al. [19], [20] identified which predictor variables are best in identifying RLR. Here, variables were considered significant if stated so by the author. Models which included common variables were compared to the number of models which identified the variable as significant. These findings of the comparison of the number of statistical models which considered a common variable and those which found the common variable significant are demonstrated in Figure 7.

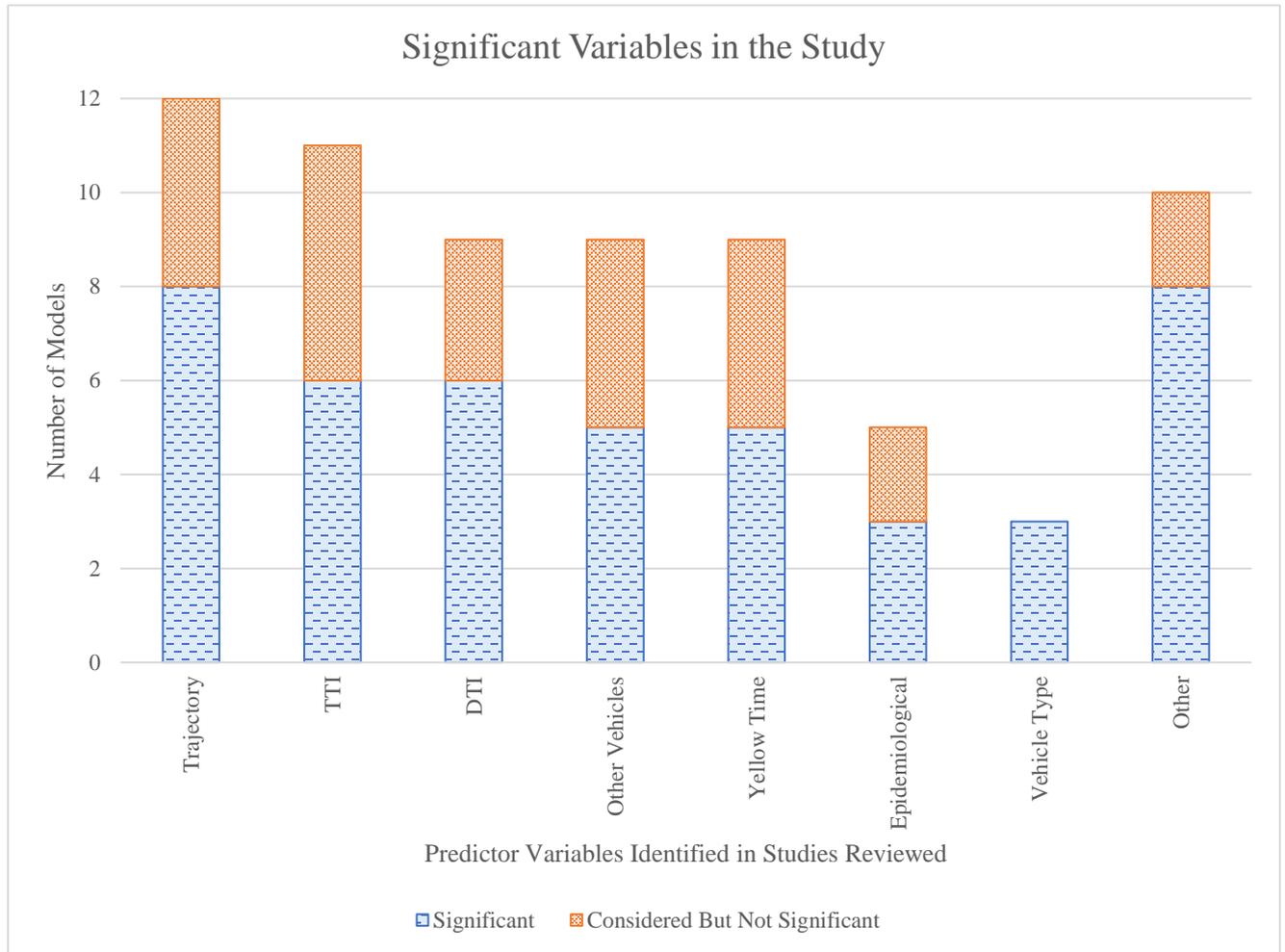

FIGURE 7. Comparison of the number of statistical models which considered a common variable (orange+blue) and those which found the common variable significant (blue).

Figure 7 indicate that TTI and yellow time remaining are highly relevant in predicting RLR and SAG behaviours. Yellow time remaining is demonstrably one of the best indicators of SAG decisions and RLR. Each statistical model has a p-value of less than 0.01 [23, 29, 33]. Ding, et al. [29] extended the notion of statistical significance by performing gradient boosting and feature importance rankings, showing that yellow time remaining ranked first, with a relative importance of 57.42%. Gates, et al. [75] used elasticity to determine the importance of the yellow time remaining and found that this value was 125.5%, more than three times higher than the second most important variable, which is platooning behaviour. The behaviour of surrounding vehicles, especially the leading vehicle, have a significant influence on a driver's SAG decision and participation in RLR behaviour [23, 29, 33]. This surrounding vehicle impact has been extended beyond vehicles within the same lane in SAG and RLR behaviour. For instance, Gates, et al. [75] found vehicles were 41% more likely to commit RLR if they were travelling as a platoon of vehicles. These findings of Gates, et al. [75] were corroborated by Ren, et al. [23], who successfully incorporated car behaviour in adjacent lanes into their prediction model, and studied in-depth by Ding, et al. [29], who found that the time gap between a vehicle and its leading vehicle was the second most important feature considered, with a relative importance of 21.12%.

Trajectory and DTI are often considered, albeit with mixed results in terms of predictor significance. Trajectory most commonly observes the speed or mean speed of a vehicle prior to the intersection [19, 20, 31-34, 75]. Some authors have extended mean speed to consider accelerations [19, 20, 23]. However, acceleration did not demonstrate any superiority to speed, and, in the case of Jahangiri, et al. [19], the maximum acceleration and decelerations were ranked lower than the velocity at the yellow onset. In the case of Elhenawy, et al. [31], speed was significant in their original logistical model but became insignificant once driver aggression was included in the model.

While it appears DTI performs poorly, it has only been found insignificant in two studies [32, 75]. One issue with this DTI measure is that it is similar to TTI in that it attempts to gauge RLR and SAG behaviour without considering variable speed limits. The closer the DTI is measured to the intersection, the better the model will be at predicting these driver decisions [26]. Thus, it is difficult to discern whether DTI is an inappropriate measure or whether the measurement distance was simply too far in these studies to make it a viable predictor variable. Each study, which states variable significance, either considered both TTI and DTI significant or only considered one of the measures, thereby making it difficult to compare. The only study that performed a direct comparison of TTI and DTI is Jahangiri, et al. [19], who found the feature importance of TTI was the highest of all variables, whereas DTI was only third.

Epidemiological factors typically included were the age and gender of the participant. Elhenawy, et al. [31] found that age was significant, while gender was not. These findings corroborate with those from Jahangiri, et al. [19], who ranked gender as the 10$^{th}$ and 11$^{th}$ most important factors, whereas age ranked significantly higher at 7$^{th}$. Vehicle type is typically insignificant, as demonstrated by Che Puan, et al. [34]. While Pathivada and Perumal [33] found different vehicle types significant or insignificant, these were not grouped into a categorical variable, therefore making it difficult to discern whether this class as a whole had a significant impact on stop-go decisions.

'other' variables considered in RLR and SAG behaviour prediction are generally insignificant. Pavement condition has never been demonstrated as significant [31, 32], nor has police presence [32] or secondary task interactions [20, 32]. Drivers' aggression was found a significant parameter for stop go behaviour prediction [31] and also variables closely related to other vehicles, such as queue length and occupancy [23].

## V. COUNTERMEASURE TECHNOLOGIES

Given the large impact of driver behaviour on RLR and SAG related collisions, it is necessary to analyse current models and technology which use driver behaviours to predict their intention to engage in RLR and SAG at intersections and pedestrian crossings. Ideally, these models are affordable and, most importantly, reduce the rate of road traffic collisions, injuries, and deaths as countermeasures.

Recently, a cooperative intelligent transportation system (C-ITS) has been recognised for safety implementation to reduce RLR and onset yellow events. In this system, vehicles are facilitated with C-ITS vehicle stations (V-ITS-S) and communicate with other vehicles and infrastructure nearby regarding the prediction and warning of safety-critical situations such as advanced red light warning and warning for vulnerable road users (VRU) crossing at the signalised intersection. This technology extends driver senses by producing safety warnings. For instance, it provides safety information even when the driver has a poor line of sight and so helps maintain robust road safety at intersections and pedestrian crossings [76-78].

Similarly, the Dynamic all red extension (DARE) system is designed to prevent collisions from occurring during RLR violations [79]. The DARE system automatically maintains the red signal at all intersection legs to reduce RLR violations. It does so by extending the red interval time dynamically by predicting a driver's intent to engage in RLR. By manipulating RLR violations, this system also prevents right angle collisions [80]. Proper implementation of such technologies may improve driver behaviour prediction prior to RLR at the signalised intersection.

Though prediction is important, systems need to be in place when these models fail to predict/identify RLR. Several collision prevention systems have been identified and are discussed in the following sections as a secondary set of countermeasures against RLR.

### A. ALL RED EXTENSION AND DYNAMIC ALL RED EXTENSION (DARE)

DARE has primarily seen use in Europe, Australia, and the USA to reduce collision rates which would typically occur as a result of RLR. It is considered one of the leading right angle collision mitigating technologies for RLR vehicles. In 2011, the North Carolina transportation department implemented DARE in nine traffic signals across the state, analysing the impact of DARE on the driver's behaviour at the intersection [79]. By analysing kinematic motion and time to collision among vehicles, the all-red interval time was calculated for the vehicles running the red light after the dilemma zone [81]. If DARE is incorporated into C-ITS technology for real time implementation, RLR rates can be reduced significantly.

### B. COUNTDOWN TIMERS

Countdown timers are also being implemented on intersections worldwide. This technology displays the remaining time for the red signal to the driver. Psychologically, this comforts the driver [82] and ensures higher safety by reducing traffic entry on the red light [83-85]. However, some studies have found that countdown timers actually increase the occurrence of RLR [86, 87] because a longer countdown timer may induce driver aggression, resulting in traffic violations [14]. Drivers often overly focus on countdown timers, and it becomes a method of distraction, resulting in the driver not focusing on their environment and being unaware of potential collisions [88].

### C. ACTUATED ADVANCED WARNING SYSTEM (FLASHING LIGHT)

The actuated advance warning system has an uninterrupted traffic monitoring facility in the dilemma zone and was introduced by The Nebraska Department of Transportation in the USA to predict the yellow onset at the signalised intersection. The actual advanced warning system can notify the driver of the SAG decision via signs and flashing signals based on the prediction [89-91]. A disadvantage of this technology lies within the signal display. The advanced stop warning signal, which indicates that drivers will soon have to stop, has been confused with the need to immediately stop [92].

### D. INTELLIGENT DRIVER WARNING SYSTEM (IDWS)

The IDWS has been implemented via safety messages. In order to prevent a traffic collision, driver behaviour was predicted from both test track and motorway data. However, the models from motorway data showed some error characteristics as they considered essential input variables like controlled vehicle speed, range and range rate to the leading vehicle, and the throttle angle history. These parameters have not been considered sufficient for practical implementation of the technology, and further exploration of this technology is needed before it can be widely implemented [93].

### E. COOPERATIVE VEHICLES INFRASTRUCTURE SYSTEMS (CVIS)

CVIS has been implemented using a binary mixed-integer linear program (BMILP) to eliminate the dilemma zone issue, showing significant promise to reduce the traffic queue near the intersection. The capacity of CVIS was limited due to considering only two lanes and four dilemma zones [94]. It was proposed that more extensive testing in different environments was required to demonstrate the robustness of the BMILP.

### F. CONNECTED VEHICLE (CV)

CV technology is a widely studied area of vehicle safety that involves a semi-automated vehicle that supplies drivers with a warning or assisted information through the transmission of communication from vehicle to vehicle, vehicle to road users and infrastructure. Unlike other technologies, C.V.s are not heavily limited by proximity to the intersection and can supply the driver with essential safety information well in advance. However, the major limitation of this CV technology is that it requires all vehicles have this technology if they are to communicate between vehicles. It is, therefore, currently impractical to implement this technology commercially [95]. A recent European Union survey with 27,656 respondents identified 45% of respondents felt comfortable with the introduction of connected vehicles on the road [96]. Studies have utilised real-time data to analyse the impact of this technology on RLR using least-squares SVM learning models [97]. The promise of this technology to improve safety, particularly when combined with existing countermeasure technology, such as DARE, has garnered increasing attention recently [98]. Moreover, the advent of artificial intelligence has brought innovation, solving different problems of the connected vehicle technology that drive the connected vehicle technology to be used practically on-road [99-101] and maritime [102, 103] too.

### G. ADVANCED DRIVING ASSISTANT SYSTEM (ADAS)

ADAS has been developed to be incorporated at different levels of autonomous driving, including connected vehicles. Significant efforts have been making to improve ADAS-based algorithms within autonomous vehicles [104]. Also, innovative controllers have been designed to regulate vehicle speed and detect obstacles in front of the vehicle [105]. In order to avoid collisions due to lane changing manoeuvres, human-machine interactions are implemented using a predictive control framework based on corrective torque optimisation [106]. The American Automobile Association recommends that if driver awareness can be improved using ADAS, doing so might save 2.7 million crashes, 1.1 million injuries, and 9,500 deaths per year [107]. A similar technology, Adaptive Cruise Control (ACC), can improve safety and be commercially viable by incorporating an onboard radar and camera to identify the primary target and regulate the vehicle speed limit. ACC systems not only ease driver effort but also prevent crashes. However, the acc system is limited in identifying target vehicles in neighbouring lanes and needs extensive research before it can be widely implemented [108].

## VI. DISCUSSION ON COUNTERMEASURE TECHNOLOGIES

Among the defined countermeasure technologies in this study, DARE is most effectively used to prevent traffic collisions as probable aftermath of red light violation. Usually, when the red signal is activated for any direction of the intersection to stop vehicles passage on that route, the right angle direction of the intersection is activated with a green signal to allow vehicle passage in a perpendicular direction of the stopped route. When a vehicle violates the red signal, it illegally appears in the intersection conflict zone. Meanwhile, the perpendicular side of the intersection is activated green signal, allowing right angle vehicles to appear at the intersection conflict zone as well that leads to right angle collision among a green signal vehicle and the red light runner. In order to remedy that deadly traffic circumstance, all red extension technology is imposed where the red signal is activated in all directions at the intersection for any predicted red light violation. This extends the red signal of the vehicles waiting at the right angle direction, delaying their green signal activation time. Such a formula improve road safety preventing deadly traffic collision but, at a time, imposes strong traffic fine ticket to the red light violating driver for violating traffic rules. Thus, all red extensions improve road safety. The DARE system is further improved and economical where the extension of red signal activation time for right angle vehicles are measured dynamically based on the time required for the red light runner to pass the conflict zone of the intersection. Moreover, the red light violation is prevented more precisely by incorporating the DARE system with CV technology where vehicles wirelessly communicate with other C.V.s and road infrastructures. Vehicle kinematic information and GPS position are easily tracked in CV technology using the embedded devices in vehicles and road infrastructures other than external cameras, and road signals can also operate dynamically, activating all red for any detected red light violation incident. With statistical and machine learning approaches (reviewed and discussed in section II) considering red-light running incidents as datasets, drivers' intend to violate the red signal is predicted earlier. This helps DARE and CV technology to get confirmed of the red light runner in advance, and they can take more accurate subsequent countermeasures using that time advantage.

Among the other countermeasures, countdown timers are also effective, but drivers often get more aggressive for extended countdown timers and rush to pass the conflict zone. The flashing light in the road signal confuses the drivers, especially in the dilemma zone. Drivers tend to stop immediately on the activation of the flashing light while they still have enough time to pass the intersection conflict zone. However, the following vehicle gets confused about the unexpected stopping behaviour of the reading vehicle that leads to the rear end collision in the dilemma zone. Considerably, ADAS is much reliable to prevent any type of collisions, but it is very expensive to install ADAS in all vehicles individually.

## VII. RESEARCH GAPS AND RECOMMENDATIONS

Some research gaps on RLR behaviour prediction using connected vehicle technology are discussed below:

- Predicting driver RLR behaviour prediction within C.V.s and issuing appropriate warnings is complex, with limited research going beyond the use of traditional machine learning and shallow artificial neural network approaches to create prediction models. Deep neural networks represent an excellent opportunity to advance the understanding of these RLR behaviours with accurate predictions and can be achieved by analysing naturalistic driving datasets. Then the accuracy and robustness of soft machine learning and shallow artificial neural network models can increase if they are trained on a broad range of different contexts. If this data includes trajectory information at signalised information that changes with time sequence order, then time series prediction of RLR behaviour using recurrent neural network models can be used as a highly accurate prediction method, even more so than deep neural network models. The development and implementation of these models is the next step in improving RLR behaviour prediction modelling.

- Combining DARE countermeasure technology with C-ITS has the potential to predict RLR behaviour. However, the robustness of the technology needs future improvement. Though countdown timers at intersections are welcoming and easy to implement, researchers need to understand why this technology sometimes increases red light violations [87]. It is necessary to implement variations of this technology were, such as internal modifications or implementation with C-ITS, so that safety can be further improved. Studies into the distraction of countdown timers are also necessary so that these increases in red light violations can be mitigated.

- Increasing features for model input improve the performance of RLR behaviour prediction models. Intersection geometry, surrounding land use, day and night period difference, peak hour, signal visibility, and emotional state of the driver are some of the many additional parameters which may lead to better prediction accuracies of RLR prediction models.

- Though research has primarily focused on heterogeneous traffic conditions with conventional vehicles for RLR behaviour prediction, mixed traffic conditions should be taken into consideration for RLR behaviour prediction, including conventional vehicles with connected vehicles and autonomous vehicles.

- Implementation of more precise measuring technologies such as using EEG signals based sensors for RLR behaviour prediction is a recent invention, but it needs major updates in sensor technologies to be used feasibly and presents a good balance between high data quality and participant comfort [8].

Eliminating trajectory and GPS errors as model inputs [109] must improve the data noises and scenario labelling, which help in increasing RLR behaviour prediction using supervised learning machine learning algorithms.

## VIII. CONCLUSION

Traffic violations and collisions for RLR and improper turning take a toll on human life and property. Millions in budget and countless efforts have been used to improve road safety; however, traffic collisions which result from RLR are still too high. Evidently, RLR is primarily caused by the intentional act to engage in RLR behaviour, and the unintentional act caused by the SAG decision within the dilemma zone. The review of SAG and RLR prediction models, their parameters and data collection methods confirmed that the time remaining on the yellow light, the presence of other vehicles, and TTI are often statistically significant parameters used to identify RLR and SAG. Furthermore, A.I. and machine learning algorithms were found to be a highly accurate alternative to the usually employed binary logistic regression model. Countermeasure technologies, such as DARE, have proven a popular choice for reducing RLR. It was found that the incorporation of this technology with C-ITS is very promising, and more research into this integration is encouraged. Several future research directions were also extrapolated. Among these, it is suggested that driver RLR behaviour should be analysed for C.V.s, in mixed traffic environments and with different parameters such as time of day and visibility.

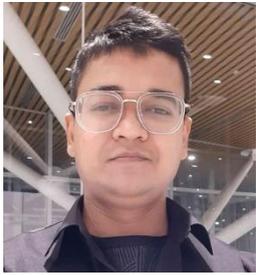

**MD MOSTAFIZUR RAHMAN KOMOL** has recently started PhD degree at Centre for Robotics in Queensland University of Technology, jointly with CSIRO Data61 Robotics and Autonomous System Group. He completed his Master of Philosophy in Intelligent Transportation System Research at Queensland University of Technology (QUT), Australia with Centre for Accident Research and Road Safety – Queensland (CARRS-Q), Australia. He is also working on the Ipswich Connected Vehicle Pilot (ICVP) project, the largest connected vehicle pilot project in Australia in affiliation with the Queensland Department of Transport and Main Roads. Here, he is involved in analysing connected vehicle data, creating algorithms for automated data labelling for vehicle movements and advanced red-light warning at intersections, analysing driving behaviours and crash severity at critical factors, building prediction models for different driving behaviours using machine learning and deep learning. He is skilled at python and MATLAB coding for data analysis using different statistical and machine learning approaches. Recently, he is also working as a research officer in CARRS-Q. Here, he is involved in an observational study of bicyclists delivering food in city centers, their characteristics and conflict records with road users. He worked at the University of Queensland as a tutor on computer engineering & data science with a UQ Australia Award Scholarship. Recently, he has been selected as a finalist of ITS Australia Young Professional Award 2021.

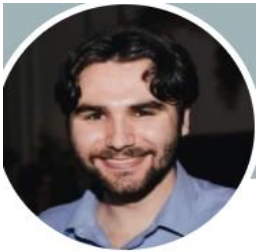

**JACK PINNOW** is an Engineer at Department of Transport and Main Roads (TMR), Australia. He worked in Ipswich Connected Vehicle Pilot project solving data analytics problems at TMR. He also worked at Research officer for QUT. He is graduated from QUT holding degrees in Mathematics (operations research) and Chemical & Process Engineering (honours). He is passionate about optimising complex, real-world systems using mathematical techniques. His interests include mineral processing, scheduling, supply chain planning and industrial chemical synthesis. His current goal is to create tools which allow for fast and accurate simulation of industrial processes. Not only will developing these tools reduce operational costs, they will also reduce operational risks.

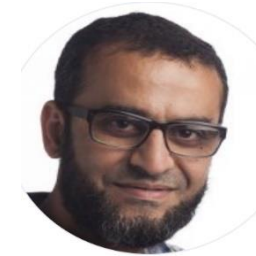

**MOHAMMED ELHENAWY** is a Senior Research Fellow at Centre for Accident Research and Road Safety Queensland and a faculty member at Queensland University of Technology. He is a PhD from Virginia Polytechnique Institute and State University under the Transportation Research Lab. He is currently working on Ipswich Connected vehicle Pilot Project and also worked on Glare on tunnel endpoints: Road safety problem, new methodological approach for analyses and simulations, Hold the Red evaluation. His research focus is on Data Science, Artificial Intelligence and Intelligent Transportation System.

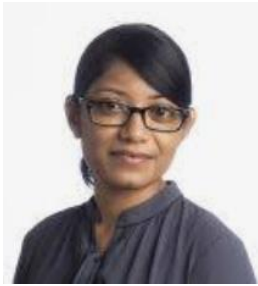

**SHAMSUNNAHAR YASMIN** is a Senior Research Fellow (Road Safety Engineer) at the Centre for Accident Research and Road Safety – Queensland (CARRS-Q), Australia. She is a Civil Engineer with ten years of research and project management experience with specialization in Transportation Engineering. She has received her Ph.D. from McGill University, M.Sc. from University of Calgary and B.Sc. from Bangladesh University of Engineering and Technology (BUET). She has previously worked as a Research Assistant in Accident Research Institute (ARI) at BUET. Before joining CARRS-Q, she was a Postdoctoral Associate and Graduate Faculty Scholar in University of Central Florida.

She has extensively worked on evidence-based and data-driven statistical analysis from the perspective of several transportation related issues. Her research expertise encompasses road safety, travel behavior, transportation planning, traffic incident management, traffic engineering, application of social media data in transportation, transit demand, sustainable urban transportation, integrated sociodemographic, built environment and land use modeling. The road safety issues that she have addressed can broadly be categorized as: (1) crash severity analysis, (2) crash frequency analysis, (3) real-time crash risk analysis and (4) road user behavior from safety perspective. In studying crash injury severity and frequency, her research deals directly with several complex methodological issues, such as those relating to population heterogeneity, endogeneity, influence of unobserved factors, data pooling and underreporting; providing new insights and evidence as to how these may be addressed in the safety field. Specifically, she have formulated, estimated and validated applications of advanced econometric models including generalized ordered, mixed/random coefficient, latent segmentation based, copula-based, simultaneous equation, time duration,

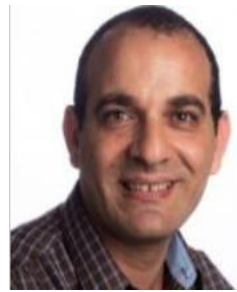

**MAHMOUD MASOUD** is a Lecturer at Centre for Accident Research and Road Safety Queensland and a faculty member at Queensland University of Technology. He is a PhD from Queensland University of Technology. He is currently working on Autonomous vehicle projects at Centre for Accident Research and Road Safety. His research focus is on Mathematical modelling, data Science, Artificial Intelligence and Autonomous Vehicle.

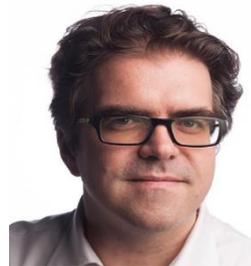

**SEBASTIEN GLASER** is Professor in Intelligent Transportation System at CARRS-Q, where he focuses on a safe and sustainable development/ deployment of Automated Driving System in interaction with others road users (drivers, cyclists, pedestrians ). He has obtained his PhD in Automatic and Control in 2004 (defining a driving assistance system in interaction with the driver) and worked, as a researcher in the development of Connected and Automated Vehicles (CAV). He was involved in several European Union initiatives (EU FP6, such as SAFESPOT on V2I communication) and in French National Research Agency (ANR) initiative (such as PARTAGE on shared control between the driver and the vehicle). Since 2009, he has worked across academic and industrial sectors and held senior researcher positions. He led the French ABV project, gathering 8 academic and industrial partners, to develop a CAV solution at low speed. He has created, with Dominique Gruyer, CIVITEC, which commercialized the research outputs on virtual environment and simulation (and is now a part of the ESI Group). He has been the deputy director and director of LIVIC (a research unit of IFSTTAR, the French institute of science and technology for transport, spatial planning, development and networks) from 2012-2015 and project leader of VEDECOM (public private partnership research institute) between 2015-2017, developing the Autonomous Vehicle prototypes. He was involved in French (ANR) and European (FP7 and H2020) initiatives on AV

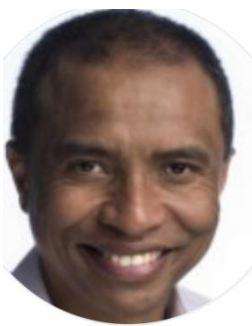

**ANDRY RAKOTONIRAINY** is the Director and Professor of Centre for Accident Research and Road Safety and founder of its Intelligent Transport Systems human factors research program, establishing its Advanced Driving Simulator laboratory. With 25 years research and management experience in computer science, he brings advanced expertise in road safety and ITS design and implementation. He is a member of the Australian Research Council (ARC) College of Experts and is a regular member of EU funded projects' advisory boards. He has authored over 250 internationally refereed papers in prestigious journals and conferences. The impact of his research is significant, with one paper cited 1034 times (4928citations in total), and he has an h-index of 34. He has been awarded 11 Australian Research Council (ARC) grants, serves on international conference and journal committees, and reviews internationally competitive grants. Professor Rakotonirainy's Intelligent Transport Systems (ITS) research has been recognised both nationally and internationally. He has proactively investigated the use of existing and emerging ITS from multiple disciplines such as computer science, mathematics, human factors, engineering, psychology and sociology. His research has made extensive use of driving simulators, traffic simulators and instrumented vehicles for developing system prototypes, assessing cost-benefits, understanding human errors and evaluating system deployment. He has been successful in securing numerous competitive grants and has established partnerships with many road safety stakeholders. Presently, he is involved in a€6,4 million EU funded (Horizon 2020) project called Levitate led by Loughborough University on 'Societal Level Impacts of Connected and Automated Vehicles'..